\documentclass{article}

\usepackage{arxiv}

\usepackage[utf8]{inputenc} % allow utf-8 input
\usepackage[T1]{fontenc}    % use 8-bit T1 fonts
\usepackage{hyperref}       % hyperlinks
\usepackage{url}            % simple URL typesetting
\usepackage{booktabs}       % professional-quality tables
\usepackage{amsfonts}       % blackboard math symbols
\usepackage{nicefrac}       % compact symbols for 1/2, etc.
\usepackage{microtype}      % microtypography
\usepackage{graphicx}
\usepackage{natbib}
\usepackage{doi}

\usepackage{algorithm}
\usepackage{algorithmic}
\usepackage{booktabs}
\usepackage{multirow} %multirow table

\title{Iterative Training: Finding Binary Weight Deep Neural Networks with Layer Binarization}

\date{} 					% Or removing it

\author{ \href{https://orcid.org/0000-0003-2830-2820}{\includegraphics[scale=0.06]{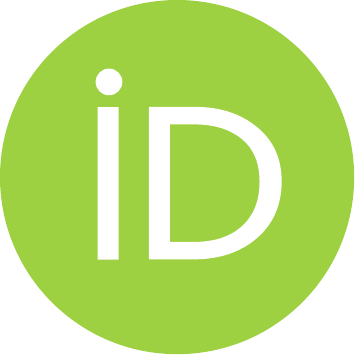}\hspace{1mm}Cheng-Chou~Lan} \\
	Rakuten Mobile, Inc.\\
	\texttt{rick.lan@rakuten.com} \\
}

\renewcommand{\headeright}{}
\renewcommand{\undertitle}{}
\hypersetup{
pdftitle={Iterative Training: Finding Binary Weight Deep Neural Networks with Layer Binarization},
pdfsubject={cs.LG, cs.CC},
pdfauthor={Cheng-Chou~Lan},
pdfkeywords={Model Compression, Deep Neural Network Algorithms},
}

\begin{document}
\maketitle

\begin{abstract}

In low-latency or mobile applications, lower computation complexity, lower memory footprint and better energy efficiency are desired.
Many prior works address this need by removing redundant parameters.
Parameter quantization replaces floating-point arithmetic with lower precision fixed-point arithmetic, further reducing complexity.

Typical training of quantized weight neural networks starts from fully quantized weights.
Quantization creates random noise.
As a way to compensate for this noise, during training, we propose to quantize some weights while keeping others in floating-point precision.
A deep neural network has many layers.
To arrive at a fully quantized weight network, we start from one quantized layer and then quantize more and more layers.
We show that the order of layer quantization affects accuracies.
Order count is large for deep neural networks.
A sensitivity pre-training is proposed to guide the layer quantization order.

Recent work in weight binarization replaces weight-input matrix multiplication with additions.
We apply the proposed iterative training to weight binarization.
Our experiments cover fully connected and convolutional networks on MNIST, CIFAR-10 and ImageNet datasets.
We show empirically that, starting from partial binary weights instead of from fully binary ones, training reaches fully binary weight networks with better accuracies for larger and deeper networks.
Layer binarization in the forward order results in better accuracies.
Guided layer binarization can further improve that.
The improvements come at a cost of longer training time.
	
\end{abstract}

% keywords can be removed

\section{Introduction}

Recent works using deep convolutional networks have been successfully applied to a large variety of computer vision tasks,
such as image recognition \citep{He2016}, object segmentation \citep{He2017} and scene segmentation \citep{Chen2018}.
These networks are large. 
For example, ResNet-152 has 60.2 million parameters \citep{Zagoruyko2016}
and requires 11.3 billion FLOPs \citep{He2016}.
A large number of parameters results in a large memory footprint.
At 32-bit floating-point precision, 229.64 MB is needed to store the ResNet-152 parameter values.

In low-latency or mobile applications, 
lower computation complexity, lower memory footprint and better energy efficiency are desired.
Many prior works address this need of lower computation complexity.
In a survey paper \cite{Cheng2018}, efficient computation of neural networks is organized into four categories: 
network pruning, low-rank decomposition, teacher-student network and network quantization.

Network pruning removes redundant parameters which are not sensitive to performance.
Low-rank decomposition uses matrix or tensor decomposition methods to reduce number of parameters.
In teacher-student network, knowledge transfer is exploited to train a smaller student network using a bigger teacher network.
In these three categories, their common theme is a reduction of number of parameters.
During forward propagation, 
one of the most computationally intensive operation in a neural network is the matrix multiplication of parameters with input. With reduced parameters, FLOPs and memory footprint reduce.
With reduced FLOPs, energy efficiency improves.

In the fore-mentioned categories, network parameters typically use floating-point precision.
In the last category, network quantization, the parameters and, for some works, all computations are quantized.
For many low-latency or mobile applications, we typically train offline and deploy pre-trained models.
Thus, the main goal is the efficiency in forward propagation.
It is desirable to compute backward propagation and parameter updates in floating-point precision.
The seminal work \cite{Courbariaux2015} matches our scope.
They quantize network weights to binary values, e.g. -1.0 and 1.0, while also keeping weight values in floating-point precision for backward propagation.
During forward propagation, instead of matrix multiplication of weights with input, the sign of these binary weights specify addition or subtraction of inputs.
Memory footprint is dramatically reduced to 1 bit per weight. Energy efficiency improves because addition is more energy efficient than multiplication \cite{Horowitz2014}.

% \cite{Li2016,Courbariaux2016,Zhou2016,Wu2018} 

Prior works in network quantization \citep{Courbariaux2015,Li2016,Courbariaux2016,Zhou2016,Wu2018}
typically start training from quantizing all weights in the network.
Quantization creates error which is the difference between the original value and its quantized value.
In other words, actual weight value, $w_{q}$, is

\begin{equation} 
  w_{q} = w - w_{error}\label{eqn:error}
\end{equation}

To reduce the impact of $w_{error}$, we hypothesize that if we quantize some weights while leaving others in floating-point precision, the latter ones would be able to compensate for the error introduced by quantization.
To reach a fully quantized network, we propose an iterative training, where we gradually quantize more and more weights. 
This raises two questions.
First, how to choose the grouping of weights to quantize together at each iteration.
Second, how to choose the quantization order across groups.
A feedforward, deep neural network has many layers. 
One natural grouping choice is one group per layer. 
For the quantization order of groups, we propose a sensitivity pre-training to choose the order.
A random order and other obvious orders are chosen as comparison.

\subsection*{Contributions.}

\begin{itemize}
  \item We propose an iterative training regime that gradually finds a full binary weight network 
  starting from an initial partial binary weight network. 
  \item We demonstrate empirically that starting from a partial binary weight network 
  result in higher accuracy than starting from a full binary weight one.
  \item We demonstrate empirically that the forward order is best, compared to other obvious orders. In addition, sensitivity pre-training can further improve that.
%  \item Code will be made public online.
  \item Code is available at \url{https://github.com/rakutentech/iterative_training}.
\end{itemize}

In the sections that follow, we describe the iterative training algorithm in detail.
Next, we present the iterative training of fully connected networks using the MNIST dataset \citep{Lecun1998}
and of convolutional neural networks using the CIFAR-10 \citep{Krizhevsky2009} and ImageNet \citep{ILSVRC15} datasets.
Then we present the sensitivity pre-training for convolution neural networks.
Finally, we discuss related work and conclusion.

\section{Iterative Training}\label{training}

\begin{algorithm}[tb]
\caption{Iterative Training}
\label{alg:layer_binary}
\textbf{Input}: Input data and label\\
\textbf{Parameter}: Number of iterations, $N$\\
\textbf{Parameter}: Number of layers, $L$\\
\textbf{Parameter}: Total number of epochs, $T$\\
\textbf{Parameter}: $BinarizationOrder$ array, length $L$\\
\textbf{Output}: A trained neural network with binary weights

\begin{algorithmic}[1] %[1] enables line numbers
\STATE $BinarizationState \leftarrow zeros(L)$.
\FOR{$j \leftarrow 1$ \TO $L$}
  \STATE $layer \leftarrow BinarizationOrder[j]$.
  \STATE $BinarizationState[layer] \leftarrow 1$.
  \FOR{$i \leftarrow 1$ \TO $N$}
    \STATE BinarizeWeights($BinarizationState$).
    \STATE ForwardPropagation().
    \STATE BackwardPropagation().
    \STATE UpdateParameters().
  \ENDFOR
\ENDFOR
\STATE $i \leftarrow L*N$
\WHILE{$i < T$}
  \STATE BinarizeWeights($BinarizationState$).
  \STATE ForwardPropagation().
  \STATE BackwardPropagation().
  \STATE UpdateParameters().
\ENDWHILE
\end{algorithmic}
\end{algorithm}

A feedforward, deep neural network has many layers, say, $L$.
We study iterative training by quantizing more and more weights layer-by-layer.
Iterative training starts from one quantized layer while all other layers are in floating-point precision.
Each iteration trains for a fixed number of epochs, $N$.
Next, we quantize the next layer and trains for another $N$ epochs.
Iterative training stops when there are no more layers to quantize.
In the case of ResNet architectures, same as the original paper,
we reduce learning rate by 10 twice and continue training.
Algorithm \ref{alg:layer_binary} illustrates the iterative training regime.
As the experiments will show, 
this regime consistently finds fully quantized network 
with better accuracies than starting from an initial fully quantized network (our baseline).

For quantization scheme, we follow weight binarization in \cite{Courbariaux2015}, 
but, for simplicity, without "tricks": no weight clipping and no learning rate scaling.
In addition, we use softmax instead of square hinge loss.
The inner for-loop in Algorithm \ref{alg:layer_binary} is the same as the training regime in \cite{Courbariaux2015},
except that a state variable is introduced to control whether a layer needs binarization or not.
We use the PyTorch framework \cite{Pytorch2019}. ImageNet results in the biggest GPU memory needs and longest training time,
which are about 10 GB and about one day to train one model on a Nvidia V100, respectively.

% \usepackage{booktabs}
% \usepackage{multirow}
% \usepackage{graphicx}
% \fontsize{9pt}{10.8pt} \selectfont
\begin{table*}[tbp]
  \centering
  \resizebox{\textwidth}{!}{%
  \begin{tabular}{@{}lllllll@{}}
  \toprule
  Network                        & 300-100-10            & 784-784-10            & Vgg-5                         & Vgg-9                 & ResNet-20           & ResNet-21 \\ \midrule
  \multirow{3}{*}{Convolutional layers} & \multirow{3}{*}{} & \multirow{3}{*}{}  & \multirow{3}{*}{64, 64}       & 64, 64                & 16, 3x[16, 16]      & 64, 4x[64] \\
                                 &                       &                       &                               & 128, 128              & 3x[32, 32]          & 5x[128], 5x[256] \\
                                 &                       &                       &                               & 256, 256              & 3x[64, 64]          & 5x[512] \\
  Fully connected layers         & 300, 100, 10          & 784, 784, 10          & 256, 256, 10                  & 256, 256, 10          & 10                  & 1000 \\ \midrule
  Dataset                        & MNIST                 & MNIST                 & CIFAR-10                      & CIFAR-10              & CIFAR-10            & ImageNet 2012 \\
  Train / Validation / Test      & 55K / 5K / 10K        & 55K / 5K / 10K        & 45K / 5K / 10K                & 45K / 5K / 10K        & 45K / 5K / 10K      & 1.2M / 0 / 50K \\
  Batch size                     & 100                   & 100                   & 100                           & 100                   & 128                 & 256 \\
  \multirow{3}{*}{Optimizer}     & \multirow{3}{*}{Adam} & \multirow{3}{*}{Adam} & \multirow{3}{*}{Adam}         & \multirow{3}{*}{Adam} & SGD                 & SGD \\
                                 &                       &                       &                               &                       & Momentum 0.9        & Momentum 0.9 \\
                                 &                       &                       &                               &                       & Weight decay 1e-4   & Weight decay 1e-4 \\
  Pre-training epochs, $K$       & 150                   & 150                   & 200                           & 450                   & 300                 & 20 \\
  Epochs per layer               & 150                   & 150                   & 150                           & 150                   & 50                  & 2 \\
  Layers, $L$                    & 3                     & 3                     & 5                             & 9                     & 20                  & 21 \\ 
  Total epochs, $T$              & 450                   & 450                   & 750                           & 1350                  & 1200                & 67 \\ \midrule
  Order count                    & 6                     & 6                     & 120                           & 362,880               & 2e+18               & 5e+19 \\ 
  Weight count                   & 266,200               & 1,237,152             & 4,300,992                     & 2,261,184             & 268,336             & About 11e6 \\ \bottomrule
  \end{tabular}%
  }
  \caption{Summary of network architectures and their hyper-parameters.}
  \label{tab:params}
\end{table*}

As shown by order count in Table \ref{tab:params}, 
there is a large number of layer binarization order for a deep neural network.
In this work, we experiment with random and obvious orders, 
to show that starting from a partially quantized weight network
is better than starting from fully quantized one.
In a later section, we introduce the proposed sensitivity pre-training to select a layer binarization order.

For obvious orders, we experiment with 
the forward order, i.e., quantizing layer-by-layer from input layer towards output layer
and the reverse order, i.e., from output layer towards input layer.
We then compare to training when:
(1) all weights are quantized from start (baseline)
(2) all weights are in floating-point precision and stay so.
As the experiments will show, for bigger and deeper networks,
the forward order consistently finds fully quantized network
with better accuracies than other orders.

In the following subsections, we discuss experimental results for fully connected and convolutional networks.

\subsection{Iterative Training for Fully Connected Networks}\label{training_fcn}
  
We investigate iterative training of fully connected networks with the MNIST dataset,
which has 60,000 training and 10,000 test images. 
We use the last 5,000 images from the training set for validation 
and the remaining 55,000 as training images for all MNIST experiments.
We use no data augmentation.
We use batch normalization \citep{Ioffe2015}, no drop-out and weight initialization as \cite{He2015}.
We use softmax as classifier.

For iterative training, we train for 150 epochs per layer.
For each network architecture, the total number of training epochs is number of layers multiplied by 150 epochs.
Because there are three layers for the chosen networks, all MNIST experiments are trained for 450 epochs.
For all cases, we find the best learning rate from the best error on the validation set.
For layer-by-layer binarization cases, the best error is selected from epochs when all layers are binarized.
We then use each corresponding best learning rate for the error on the test set.
We vary the seed for 5 training sessions and report the learning curves of the average test errors in the figures.
Table \ref{tab:params} reports other hyper-parameters.

\iffalse
% TODO remove
\begin{figure}[t]
  \centering
  \includegraphics[width=0.5\columnwidth]{gfx_300_bigger_simpler_sns}
  \caption{Test errors for 300-100-10 network.}
  \label{fig:error_of_300}
\end{figure}

\begin{figure}[t]
  \centering
  \includegraphics[width=0.5\columnwidth]{gfx_784_bigger_simpler_sns}
  \caption{Test errors for 784-784-10 network.}
  \label{fig:error_of_784}
\end{figure}
\fi

\begin{figure}[t]
  \centering
  \includegraphics[width=0.5\columnwidth]{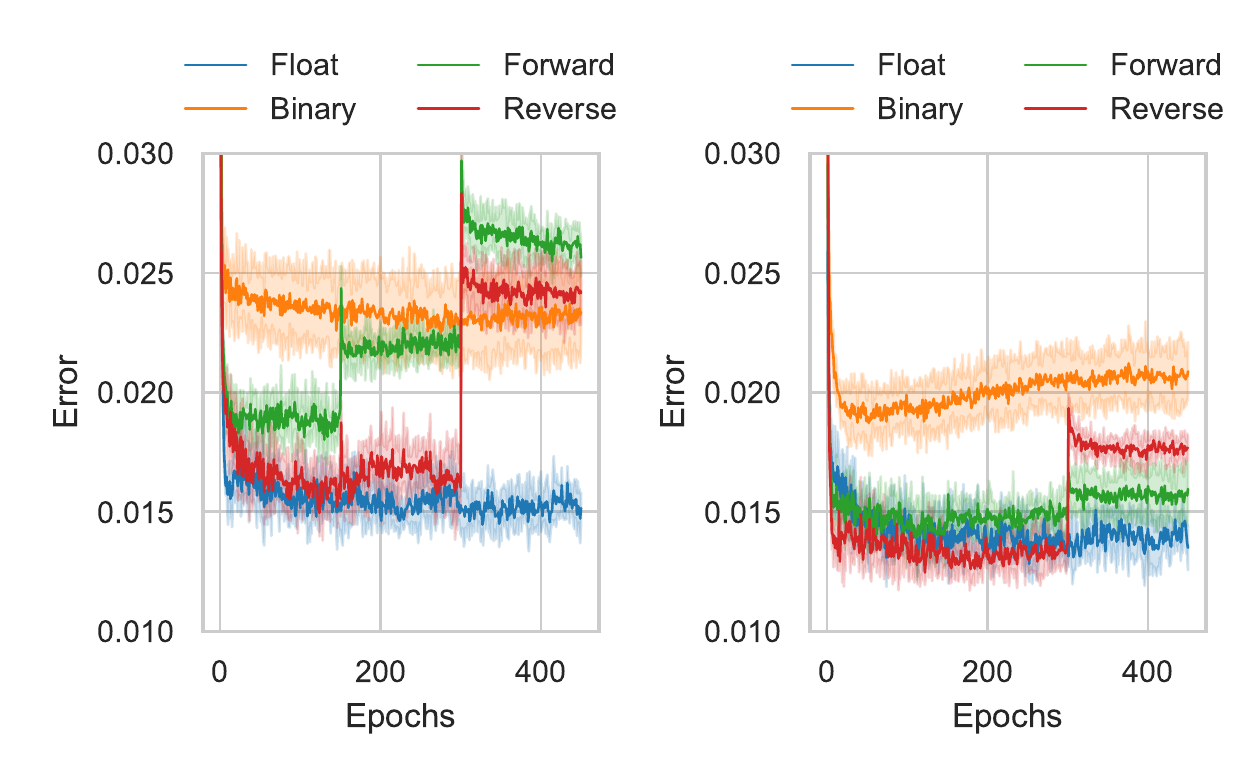}
  \caption{
    Left: Test errors for 300-100-10 network.
    Right: Test errors for 784-784-10 network.
  }
  \label{fig:error_of_300_784}
\end{figure}

% \usepackage{booktabs}
% \usepackage{graphicx}
% \resizebox{\textwidth}{!}{%
% \resizebox{0.5\linewidth}{!}{%
\begin{table}[tbp]
  \centering
  \begin{tabular}{@{}llll@{}}
  \toprule
  Case    & 300-100-10 & 784-784-10 & Improvement \\ \midrule
  Binary  & 0.023      & 0.021      & 0.002       \\
  Float   & 0.015      & 0.014      & 0.001       \\
  Reverse & 0.024      & 0.018      & 0.006       \\
  Forward & 0.026      & 0.016      & 0.010       \\ \bottomrule
  \end{tabular}
  \caption{Error improvement from using 784-784-10 over 300-100-10.}
  \label{tab:784_advantage}
\end{table}

For network architectures, we study the 300-100-10 network \citep{Lecun1998} and a bigger variant, 784-784-10.
Figure \ref{fig:error_of_300_784} shows test errors for the 300-100-10 and 784-784-10 networks.
The float case is training where all weights are in floating-point precision and stay so.
The binary case (baseline) is training where all weights are binarized from the start.
The forward case is training where layer binarization is in the forward order,
the reverse in the reverse order.
The solid lines are the mean across multiple runs and the matching shaded color is one standard deviation.

For the smaller network, 300-100-10, the binary case reaches a lower error than forward and reverse orders.
Next best is the reverse order then the forward one.
This shows that order of layer binarization matters for accuracy.
On the contrary, for the bigger network, 784-784-10, the forward and reverse cases does better than the binary one.
Binarization operation is not differentiable.
According to Equation \ref{eqn:error}, it injects a random error signal into the network.
During iterative training, some of the weights are in floating-point precision.
We hypothesize that they are compensating for the random error.
At the same time, we think bigger networks are more robust due to more parameters.

The error improvement of upgrading to a bigger network is given in Table \ref{tab:784_advantage}.
The forward and reverse orders have significantly higher improvement than float and binary, 
showing that iterative training is beneficial.
In addition, the forward order has a higher improvement than reverse.
We observe the same pattern for subsequent network architectures.
Namely, for bigger and deeper networks, 
starting from partial binary weight network, instead of full binary weight network, 
iterative training with forward weight quantization order finds full binary weight network with higher accuracies.

\subsection{Iterative Training for Convolutional Networks}\label{training_cnn}

We investigate iterative training of convolutional networks with the CIFAR-10 dataset,
which has 50,000 training and 10,000 test images.
We randomly choose 5,000 images from the training set as the validation set 
and the remaining 45,000 as training images for all CIFAR-10 experiments.
We use the same data augmentation as \cite{He2016}:
4 pixels are padded on each side, and a 32x32 crop is randomly sampled from the padded image or its horizontal flip.
We use batch normalization, no drop-out and weight initialization as \cite{He2015}.
We use softmax as classifier.

We experiment with VGG \citep{Simonyan2015} and ResNet \citep{He2016} architectures.
For iterative training of VGG architectures, we train for 150 epochs per layer.
For iterative training of ResNet-20 architecture, we train for 50 epochs per layer.
Same as the original paper, we reduce learning rate by a factor 10 twice,
once at 1000 epochs and a second time at 1100 epochs.
Then stop training at 1200 epochs.
Using same methodology as MNIST experiments, 
for all cases, 
we use the validation set to tune the learning rate
and test set to report errors.
Table \ref{tab:params} reports other hyper-parameters. 

\iffalse
% TODO remove
\begin{figure}[t]
  \centering
  \includegraphics[width=0.5\columnwidth]{gfx_vggconv2final_sns}
  \caption{
  Test errors for Vgg-5 network.
  }
  \label{fig:error_of_conv2}
\end{figure}

\begin{figure}[t]
  \centering
  \includegraphics[width=0.5\columnwidth]{gfx_vggconv6final_sns}
  \caption{
  Test errors for Vgg-9 network.
  }
  \label{fig:error_of_conv6}
\end{figure}
\fi

\begin{figure}[t]
  \centering
  \includegraphics[width=0.5\columnwidth]{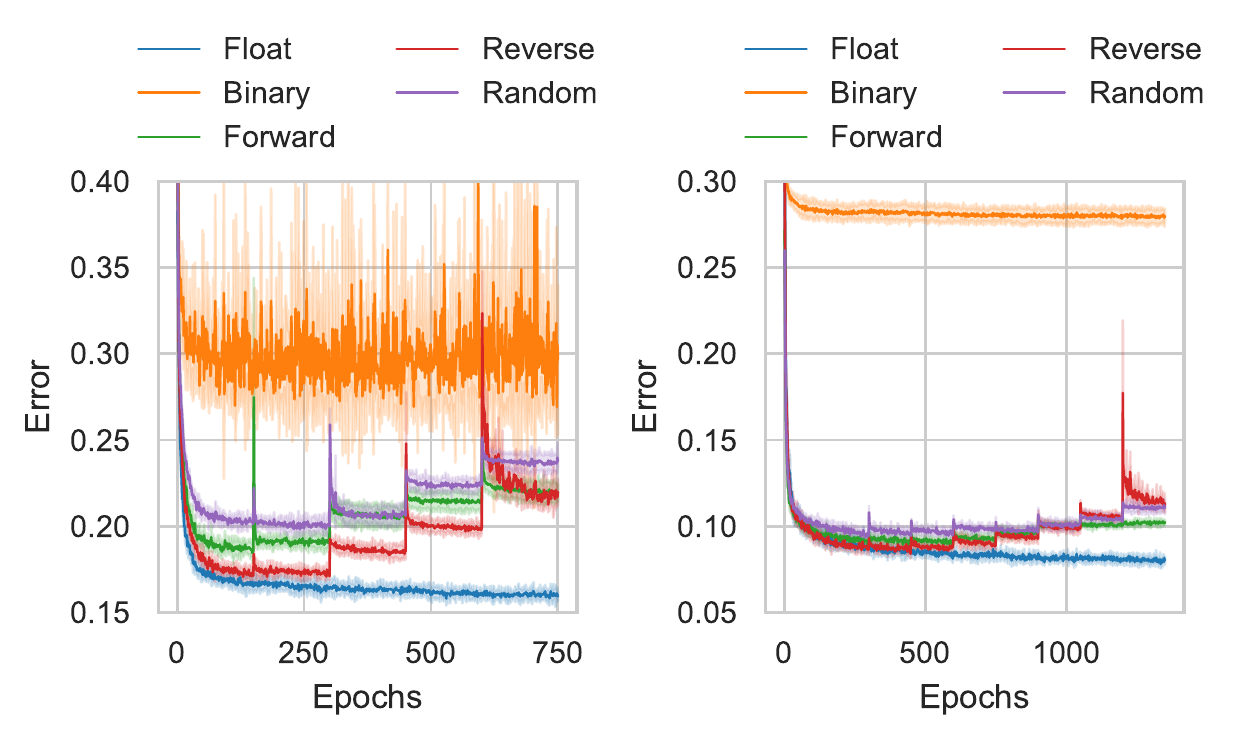}
  \caption{
  Left: Test errors for Vgg-5 network.
  Right: Test errors for Vgg-9 network.
  }
  \label{fig:error_of_conv2_conv6}
\end{figure}

For VGG architecture, we study a shallower, Vgg-5, and a deeper network, Vgg-9.
As their names suggest, Vgg-5 has 5 layers and Vgg-9, 9.
Figure \ref{fig:error_of_conv2_conv6} shows test errors for Vgg-5 and Vgg-9 networks.
The float case is training where all weights are in floating-point precision and stay so.
The binary case (baseline) is training where all weights are binarized from the start.
The forward case is training where layer binarization is in the forward order,
the reverse in the reverse order
and the random case, a randomly selected order.

For both network architectures, 
the binary case has the highest error and the float case the lowest error.
In the same pattern as the larger MNIST network,
starting from partial binary weight networks, 
iterative training finds full binary weight networks that have lower error than the binary cases.
For Vgg-5, a shallower network, the ascending error ranking is reverse, forward then random.
For Vgg-9, a deeper network, the ranking is forward, random then reverse.
This shows again that layer binarization order matters.

% \usepackage{booktabs}
% \usepackage{graphicx}
% \resizebox{\textwidth}{!}{%
% \resizebox{0.5\linewidth}{!}{%
\begin{table}[tbp]
  \centering
  \begin{tabular}{@{}llll@{}}
  \toprule
  Case   & Vgg-5 & Vgg-9  & Improvement \\ \midrule
  Binary  & 0.30  & 0.28   & 0.02        \\
  Float   & 0.16  & 0.08   & 0.08        \\
  Reverse & 0.22  & 0.1126 & 0.1074      \\
  Forward & 0.22  & 0.1025 & 0.1175      \\ \bottomrule
  \end{tabular}
  \caption{Error improvement from using Vgg-9 over Vgg-5.}
  \label{tab:vgg9_advantage}
\end{table}

The error improvement of upgrading to Vgg-9 from Vgg-5 is summarized in Table \ref{tab:vgg9_advantage}.
There is a small improvement for the binary case.
The float case has a significantly higher improvement than binary.
Next higher is the reverse case.
Finally, the forward case has the highest improvement.
In the same pattern as in the MNIST experiments, favoring iterative training and the forward order.

As shown in Table \ref{tab:params}, although Vgg-9 has a smaller number of weight parameters than Vgg-5,
it has more layers.
Iterative training continues to be beneficial.
We hypothesize that this is due to a more gradual rate of total binarization.
For Vgg-9, as each layer is binarized, 
relatively more weights stay in floating-point precision to compensate for the random noise injected by the binarization operation.

\begin{figure}[t]
  \centering
  \includegraphics[width=0.5\columnwidth]{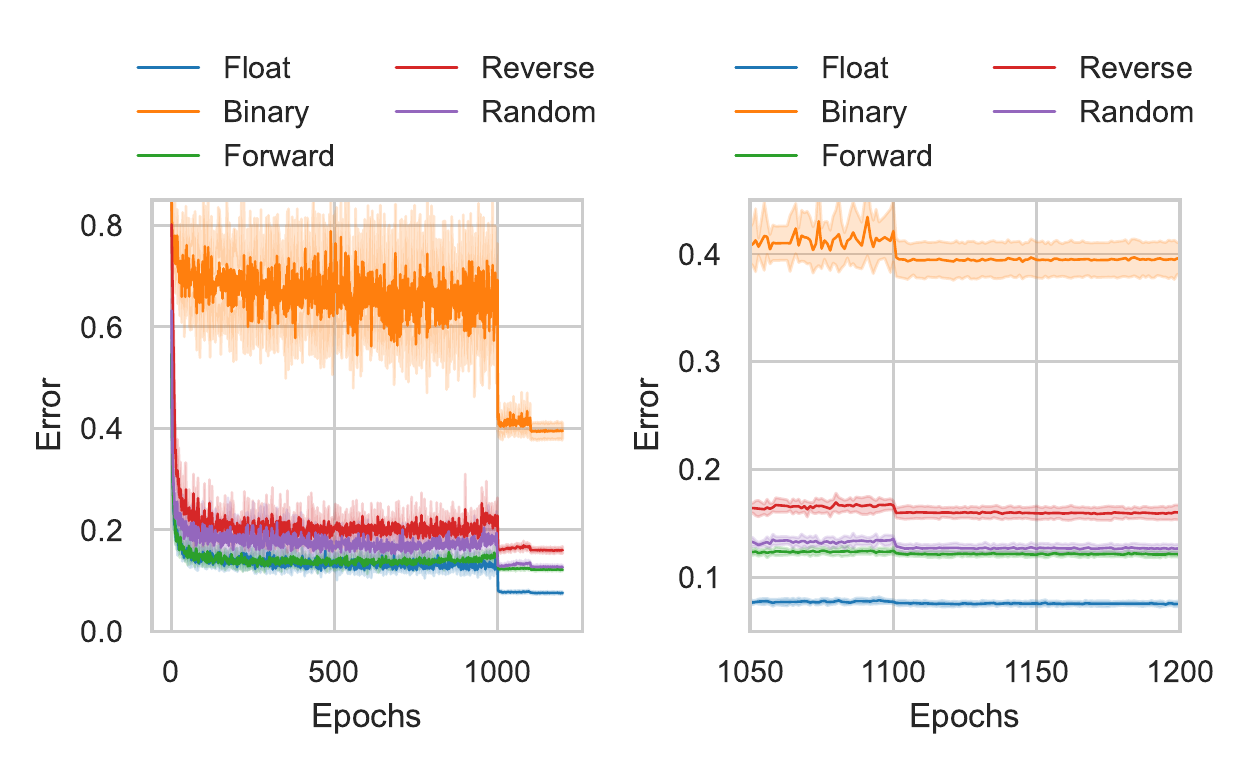}
  \caption{
  Left: Test errors for ResNet-20 network. 
  Right: Zoom to final epochs.
  Learning rate is reduced by 10x at 1000 and again at 1100 epochs.
  }
  \label{fig:error_of_resnet20}
\end{figure}

For an even deeper network, we study ResNet-20 from \cite{He2016}, which has 20 layers, as its name suggests.
Figure \ref{fig:error_of_resnet20} shows test errors for the ResNet-20 network.
The binary case has the highest error and the float case has the lowest error.
In the same pattern as other network architectures, 
iterative training finds full binary weight networks that have lower error than the binary case.
In increasing error order is forward, random and reverse.
Again, showing that the order of binarization matter and the forward order has advantage.
In the next section, we propose a sensitivity pre-training to select a binarization order.

\section{Sensitivity Pre-training}\label{sensitivity_pretraining}

In prior sections we demonstrated empirically that starting from a partial binary weight network 
results in higher accuracy than starting from a fully binary weight one for larger and deeper networks. 
In this section, we describe the proposed sensitivity pre-training to choose a the binarization order.

For shallower neural networks like the 3-layer fully connected networks for the MNIST dataset,
exhaustive search for the best binarization order is possible.
For deeper neural networks such as Vgg-5, Vgg-9 and Resnet-20, it is impractical to do so, as shown by order count in Table \ref{tab:params}.
However, we can obtain a measure of error sensitivity to layer quantization.
Then let the sensitivity be a guide for binarization ordering.

Sensitivity is computed as follows.
We train $L$ models, where in each model only the weights of the L-th layer is binarized while others are in floating-point precision.
We train for $K$ epochs and, as before, use validation set to tune the learning rate to get the best validation error for each sensitivity model. 
$K$ for Vgg-5 is 200 and for Vgg-9 is 450.
$K$ for ResNet-20 is 300.
For ResNet, same as the original paper, we reduce learning rate by 10 twice, one at epoch 200 and again at epoch 250.
Then we rank these $L$ best validation errors in ascending order.
This becomes the ascending layer binarization order for iterative training.

\iffalse
\begin{figure}[t]
  \centering
  \includegraphics[width=0.5\columnwidth]{gfx_vggconv2final_sensitivity_sns}
  \caption{
  Test errors for Vgg-5 network.
  Ascending and descending orders are chosen by sensitivity pre-training.
  }
  \label{fig:error_of_conv2_with_sensitivity}
\end{figure}

\begin{figure}[t]
  \centering
  \includegraphics[width=0.5\columnwidth]{gfx_vggconv6final_sensitivity_sns}
  \caption{
  Test errors for Vgg-9 network.
  Ascending and descending orders are chosen by sensitivity pre-training.
  }
  \label{fig:error_of_conv6_with_sensitivity}
\end{figure}
\fi

\begin{figure}[t]
  \centering
  \includegraphics[width=0.5\columnwidth]{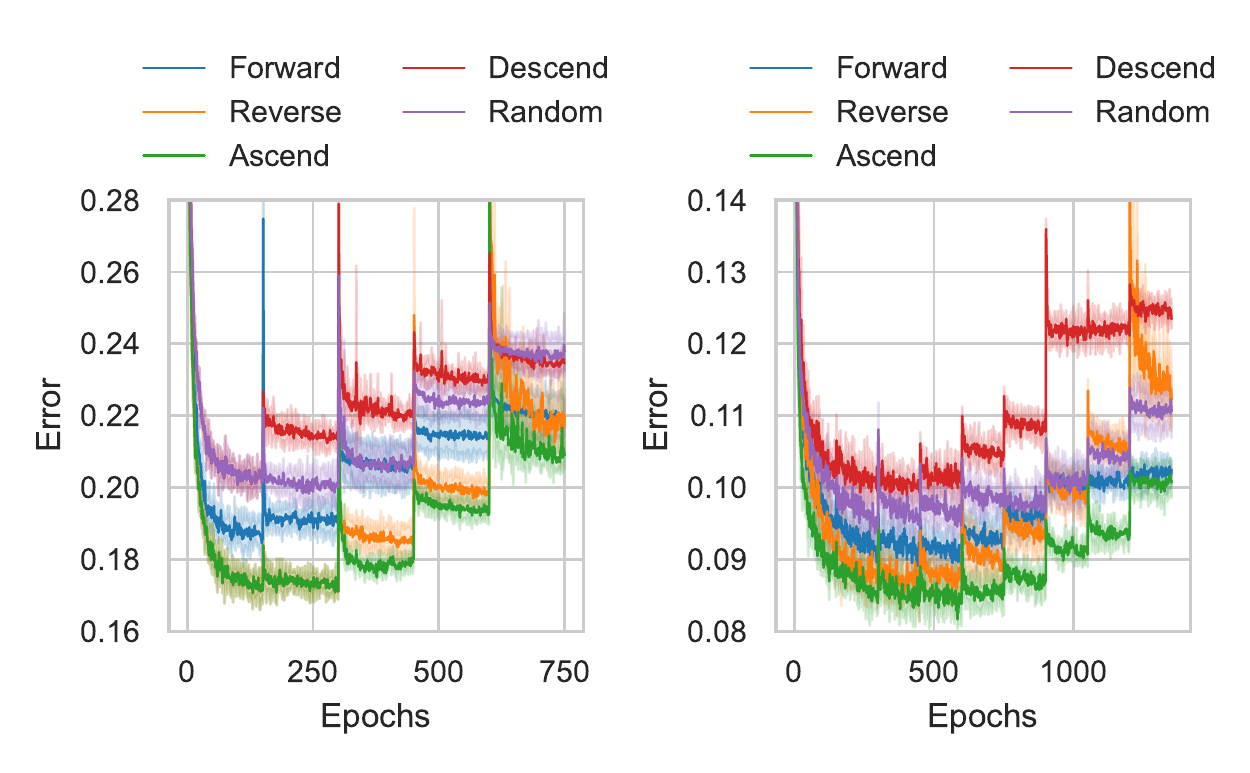}
  \caption{
    Left: Test errors for Vgg-5 network.
    Right: Test errors for Vgg-9 network.
    Ascending and descending orders are chosen by sensitivity pre-training.
  }
  \label{fig:error_of_conv2_conv6_with_sensitivity}
\end{figure}

During iterative training using ascending order, the layer that had the lowest error will be binarized first, 
while the layer that had the highest error last, meaning the latter stays in floating-point precision the longest during training.
As shown in Figure \ref{fig:error_of_conv2_conv6_with_sensitivity} for Vgg-5 and Vgg-9,
the ascending order results in a fully binary weight network with the lowest error,
beating the forward ones.
Also shown is the descending order, which is the reverse of the ascending one.
For both networks, the descending order results in error higher than ascending,
showing again that binarization order matters.
In the case of Vgg-5, the random order is worst while descending one follows closely behind.
In the case of Vgg-9, the descending one is the worst of all.
In short, the lower the error for one order, the higher its reverse order would be.

\begin{figure}[t]
  \centering
  \includegraphics[width=0.5\columnwidth]{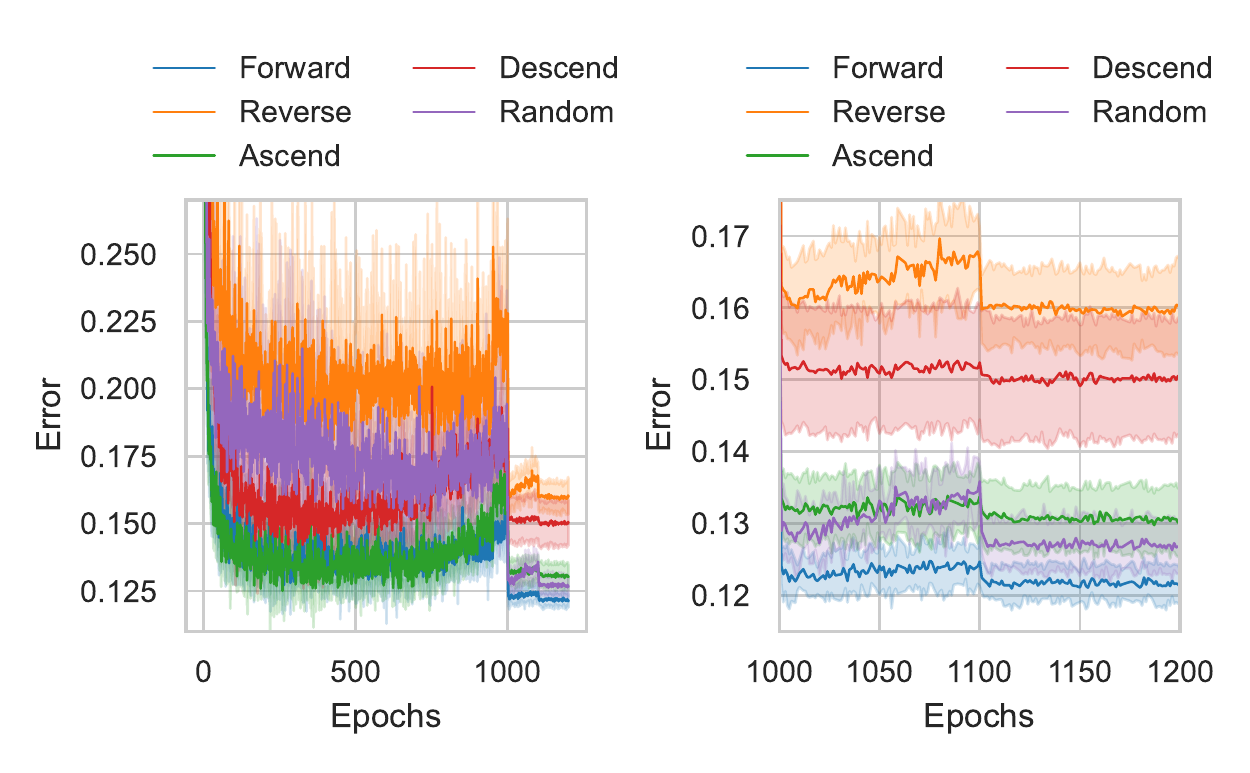}
  \caption{
  Left: Test errors for Resnet-20 network.
  Right: Zoom to final epochs.
  Ascending and descending orders are chosen by sensitivity pre-training.
  }
  \label{fig:error_of_resnet20_with_sensitivity}
\end{figure}

For ResNet-20, Figure \ref{fig:error_of_resnet20_with_sensitivity} shows the test errors with ascending and descending orders.
Unlike for Vgg-5 and Vgg-9, the forward order reaches accuracy better than both ascending and descending orders.
The proposed sensitivity pre-training considers binarization of layers independently.
We hypothesize that there may be interactions between multiple layers.

\begin{figure}[t]
  \centering
  \includegraphics[width=0.5\columnwidth]{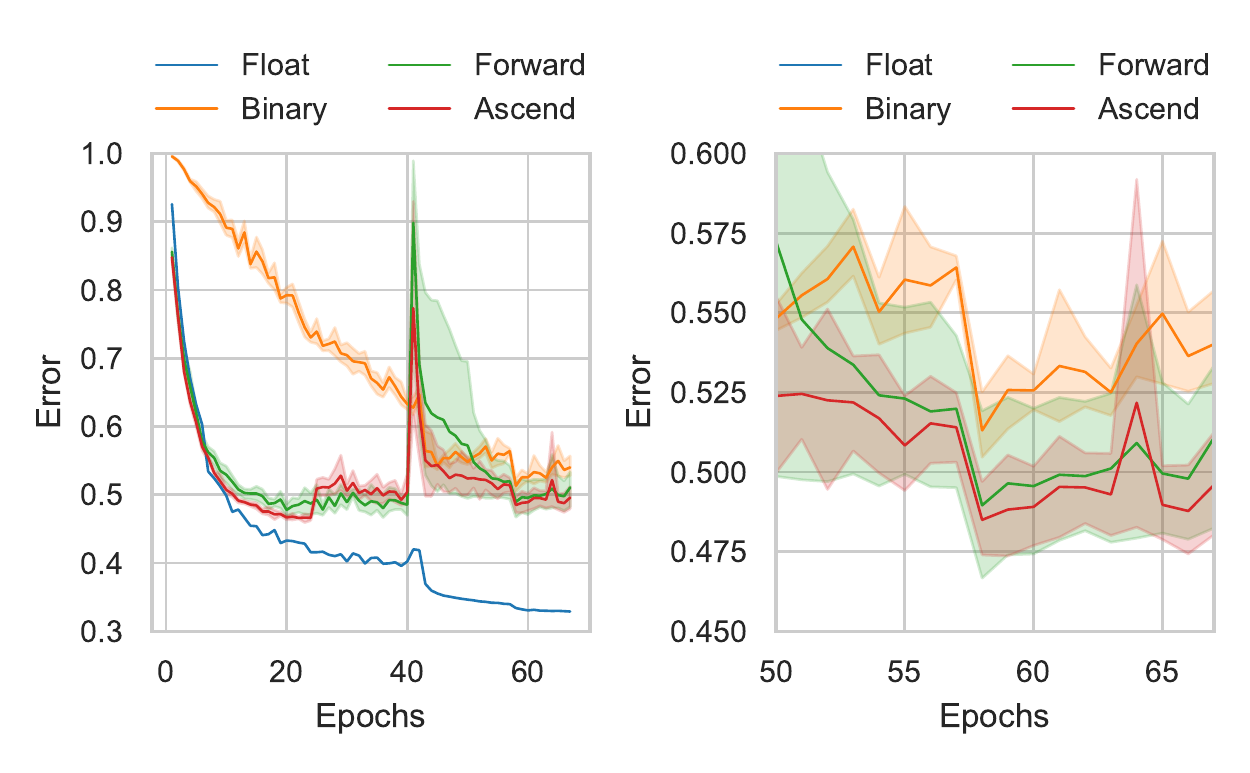}
  \caption{
  Left: Test errors for ResNet-21 network.
  Right: Zoom to final epochs.
  Ascending order are chosen by sensitivity pre-training.
  }
  \label{fig:error_of_resnet21_with_sensitivity}
\end{figure}

For ImageNet, we experiment with ResNet-18 \cite{He2016}.
Since it has 21 layers, we will refer to it as ResNet-21.
The optimizer is SGD with momentum 0.9 and weight decay 1e-4.
For sensitivity pre-training, $K$ is 20 epochs.
For each layer, we sweep 3 learning rates and use the last-epoch errors of the test set to choose the ascending order.
In the full training, we choose 2 epochs per layer.
The starting learning rate, 0.01, comes from the best learning rate in sensitivity pre-training.
Same as the orginal paper, we reduce learning rate by 10 twice, after 42 epochs and again after 57 epochs.
We stop training after 67 epochs.
The floating-point training is just one run, while all other binarization training are from 5 random-seeded runs.
Figure \ref{fig:error_of_resnet21_with_sensitivity} shows the test errors with forward and ascending orders.
The ascending order has a lower mean error than forward.
Both of which are better than binary.
Again, binarization order matters and ascending order is better than the forward one.

\subsection{Exhaustive Search}

\iffalse
% TODO remove
\begin{figure}[t]
  \centering
  \includegraphics[width=0.5\columnwidth]{gfx_300_sensitivity_sns}
  \caption{
    Test errors for 300-100-10 network.
    Ascending order is same as reverse and descending same as forward.
    132 means binarization order is layer 1, layer 3 then layer 2.
  }
  \label{fig:error_of_300_with_sensitivity}
\end{figure}

\begin{figure}[t]
  \centering
  \includegraphics[width=0.5\columnwidth]{gfx_784_sensitivity_sns}
  \caption{Test errors for 784-784-10 network.}
  \label{fig:error_of_784_with_sensitivity}
\end{figure}
\fi

\begin{figure}[t]
  \centering
  \includegraphics[width=0.5\columnwidth]{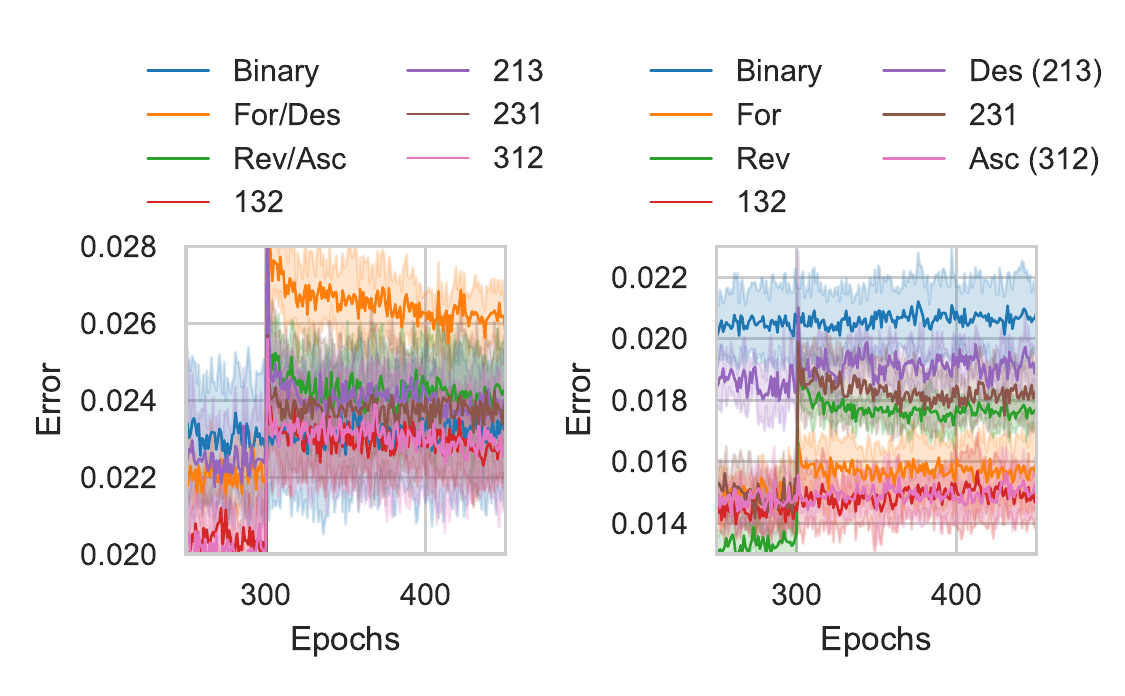}
  \caption{
    Left: Test errors for 300-100-10 network.
    Right: Test errors for 784-784-10 network.
    132 means binarization order is layer 1, layer 3 then layer 2.
    Forward order is 123.
  }
  \label{fig:error_of_300_784_with_sensitivity}
%  Old Lenet 300 caption:
%  Ascending order is same as reverse and descending same as forward.
%  132 means binarization order is layer 1, layer 3 then layer 2.

\end{figure}

For shallower neural networks like the 3-layer fully connected network for the MNIST dataset,
exhaustive search for the best binarization order is possible.
Figure \ref{fig:error_of_300_784_with_sensitivity}
shows result for all combinations of layer binarization order for 300-100-10 and 784-784-10 networks.
For the former, a smaller network, the ascending order turns out to be same as reverse.
Errors for all combinations are very close.
The best order is not the ascending one, but 132 and 312, both of which are better than binary by a small margin.
132 means binarization order is layer 1, layer 3 then layer 2.
Thus, also for 300-100-10, starting from partial weight binarization is better than from full weight binarization.

For the bigger network, 784-784-10, the ascending order is better than forward and reverse ones.
The descending order is worst of all others.
This is consistent with the results from convolutional networks.
Here, the ascending one shares with another in best accuracy.

In summary, we proposed using sensitivity pre-training as a guide for layer binarization order.
For 784-784-10, Vgg-5, Vgg-9 and ResNet-21, we have shown empircally that better accuracies are achieved.
This improvement comes at a cost of pre-training additional $L$ models.

\section{Related Work}\label{related_work}

%In a survey paper by \cite{Cheng2018}, efficient computation of neural networks is organized into four categories: 
%network pruning, low-rank decomposition, network quantization and teacher-student network.

Our work introduces an iterative layer-by-layer quantization training regime.
Although we demonstrated the results using weight binarization,
the regime is independent of quantization schemes.
We think other schemes,
e.g., \cite{Li2016} (where weights are ternary: -1.0, 0 and 1.0), may yield similar trends.
%For downside, iterative training comes at a cost of longer training time.

%However, due to nature of iterative training, training time is lengthened.

\cite{Hu2018} transforms weight binarization as a hashing problem.
Same as ours, their iterative method also operates layer-by-layer, from input layer towards output layer.
However, they start from a pre-trained network and, after weight quantization without fine-tuning, fine-tune the biases.
Ours starts from an untrained network and gradually trains a full binary weight network,
which we believe allows the network to adapt to the random noise created by the quantization operation.
In addition, their final weights are not pure binary, but power-of-2 multiples.
When constrained to pure binary, they report non-convergence.
Our iterative training does not require pure binary weights.
For future work, we can binarize using power-of-2 multiples.

\cite{Zhou2017} iterates over both pruning and quantization techniques.
First, weights are partitioned into two groups. 
Then, weights in first group are quantized to power-of-2 multiples or zero.
Next, weights in the second groups are fine-tuned, while the first group receives no parameter updates.
In the next iteration, some of weights in second group is assigned to the first group.
The process is repeated until all weights are members of the first group.
In this partitioning scheme, the first group contains weights from all layers.
It is possible to merge both methods because their partitioning is orthogonal to ours.
Once weights join the first group, their values stay unchanged for the rest of the fine-tuning.
Because our binarization is based on \citep{Courbariaux2015},
floating-point weights prior to quantization are saved for parameter updates.
Thus, during iterative training of later layers, weights of prior layer are allowed to adapt and flip signs.
However, the disadvantage is more memory are required during training.

In low-rank decompositions and teacher-student network, weights are still in floating-point precision.
For low-rank decomposition, the implementation requires decomposition operation, which is computationally expensive,
and factorization requires extensive model retraining to achieve convergence when compared to the original model \citep{Cheng2018}.
Similarly, due to iterative nature of our proposed training regime, training time is also lengthened.

\section{Conclusions and Further Work}\label{conclusion}
In this work, we proposed a simple iterative training 
that gradually trains a partial binary weight network to a full binary weight network layer-by-layer.
We showed empirically that this regime results in higher accuracy than starting training from a fully binarized weight network.
The order of layer binarization matters.
We show empircally that, for larger and deeper neural networks, 
the forward order achieves better accuracies than other binarization orders.
We proposed a sensitivity pre-training for selection of binarization order.
For 784-784-10, Vgg-5, Vgg-9 and ResNet-21, this guided order achieve better accuracies than the forward order.

Iterative training has a cost, which is lengthened training.
This trade-off may be acceptable in many applications where pre-trained models are deployed,
because efficiency in only forward propagation is needed.
A binary weight neural network dramatically reduces computation complexity, memory footprint and, thus, increases energy efficiency.
For future work, we would like to understand analytically why layer quantization works
and the optimal quantization order.

\bibliographystyle{unsrtnat}
\bibliography{paper}  %%% Uncomment this line and comment out the ``thebibliography'' section below to use the external .bib file (using bibtex) .





% \end{thebibliography}

\end{document}